# Conversational Crowdsensing: A Parallel Intelligence Powered Novel Sensing Approach

Zhengqiu Zhu, Yong Zhao, Bin Chen, Sihang Qiu*, Kai Xu, Quanjun Yin, Jincai Huang, Zhong Liu, and Fei-Yue Wang, *Fellow, IEEE*

*Abstract*—The transition from CPS-based Industry 4.0 to CPSS-based Industry 5.0 brings new requirements and opportunities to current sensing approaches, especially in light of recent progress in Chatbots and Large Language Models (LLMs). Therefore, the advancement of parallel intelligence-powered Crowdsensing Intelligence (CSI) is witnessed, which is currently advancing towards linguistic intelligence. In this paper, we propose a novel sensing paradigm, namely conversational crowdsensing, for Industry 5.0. It can alleviate workload and professional requirements of individuals and promote the organization and operation of diverse workforce, thereby facilitating faster response and wider popularization of crowdsensing systems. Specifically, we design the architecture of conversational crowdsensing to effectively organize three types of participants (biological, robotic, and digital) from diverse communities. Through three levels of effective conversation (i.e., inter-human, human-AI, and inter-AI), complex interactions and service functionalities of different workers can be achieved to accomplish various tasks across three sensing phases (i.e., requesting, scheduling, and executing). Moreover, we explore the foundational technologies for realizing conversational crowdsensing, encompassing LLM-based multi-agent systems, scenarios engineering and conversational human-AI cooperation. Finally, we present potential industrial applications of conversational crowdsensing and discuss its implications. We envision that conversations in natural language will become the primary communication channel during crowdsensing process, enabling richer information exchange and cooperative problem-solving among humans, robots, and AI.

## I. INTRODUCTION

IN recent years, intensive discussions and studies on Industry 5.0 [1]–[4] have emerged, capturing the attention of researchers, entrepreneurs, and policymakers across diverse sectors worldwide. In fact, Industry 5.0 related academic papers and technical reports can be traced back to Prof. Wang's

Manuscript received January X, 2024; revised March X, 2024; accepted May XX, 2024. This study is supported by the National Natural Science Foundation of China (62202477, 62173337, 21808181, 72071207), and Youth Independent Innovation Foundation of NUDT (ZK-2023-21). (*Corresponding author: Sihang Qiu*).

Zhengqiu Zhu, Yong Zhao, Bin Chen, Sihang Qiu, Kai Xu, Quanjun Yin, Jincai Huang, and Zhong Liu are with the College of Systems Engineering, National University of Defense Technology, Changsha 410073, Hunan Province, China. (e-mail: zhuzhengqiu12@nudt.edu.cn; zhaoyong15@nudt.edu.cn; chenbin06@nudt.edu.cn; sihangq@acm.org; xukai09@nudt.edu.cn; yin_quanjun@163.com; huangjincai@nudt.edu.cn; phillipliu@263.net).

Fei-Yue Wang is with the State Key Laboratory for Management and Control of Complex Systems, Institute of Automation, Chinese Academy of Sciences, Beijing 100190, China. (e-mail: feiyue@ieee.org).

works on Parallel Intelligence (PI) [5]–[7] and Intelligent Industries in 2000s [8]. Since then, Prof. Wang and his team have been dedicated to developing a theory with implementable frameworks and systematic processes, aimed at facilitating the realization of Industry 5.0 [9], where the core is PI. The PI framework provides a solid methodological foundation to make complex systems computable, testable, verifiable, and ultimately controllable. Consequently, by considering PI as a technical means and cyber-physical-social systems (CPSSs) as an infrastructure, the progression towards Industry 5.0 becomes feasible [10].

As data and information is crucial in industrial activities, intelligent sensing approaches [11], especially crowdsensing [12], [13] and spatial crowdsourcing [14], have been extensively employed across various stages of industrial processes (e.g., design, production, logistics, and user feedback), emerging as foundation elements of Industrial 4.0. However, in the era of Industrial 5.0, with human and social factors being introduced, the industrial systems (typical CPSSs) face some issues: excessive human involvement, lack of flexibility, and inefficient human-computer interaction [15]. Simultaneously, with the rapid advancement of the Internet of Things (IoT) and social media, coupled with the widespread use of mobile devices, vast amounts of data are being generated almost instantaneously from not only physical space but also cyber and social spaces. Consequently, traditional crowdsensing encounters challenges in adapting to the transition from CPS-based Industry 4.0 to CPSS-based Industry 5.0. Therefore, an urgent need arises for a suite of intelligent sensing schemes that can effectively capture the dynamic characteristics of Industry 5.0 across cyber, physical, and social spaces.

The recent advancement in PI has inspired researchers to propose the concept of Crowdsensing Intelligence (CSI) [16]–[19]. Through the deep integration of robotic, biological, and digital participants [20]–[23], CSI leverages their diverse sensing abilities, complementary computing resources, and cross-space collaboration to build a decentralized, self-organizing, self-learning, and continuously evolving intelligent sensing and computing space. In this space, individual skills and collective cognitive ability can be enhanced to facilitate the guidance and control of the actual space [24]. The initial exploration of CSI can be traced back to the rudiment CCEC-PTS framework [16] which integrates heterogeneous sensing resources in social transportation and collects associated data from different spaces. The formal exploration of CSI was launched by Professor Wang Fei-Yue and Chen Bin in the ongoing DHW-CSI. The previous workshops on CSI have primarily focused on the DAO-based architecture [25], [26] and fundamental components of CSI, such as participants, methods, and stages [18].

To achieve PI-based CSI, three important cornerstones are utilized. 1) The first cornerstone is the organization and operation of three types of workers: **biological (˜5%), digital (˜80%), and robotic (˜15%)**. To enable three types of workers to perform a diverse range of sensing tasks with greater autonomy and intelligence, the technology of Decentralized Autonomous Organizations and Operations (DAO) should be applied. DAO facilitates the formation of distributed, decentralized, autonomous, automated, organized, and resilient crowdsensing communities by assembling individuals who share common objectives, driven by specific mechanisms for value creation and incentive distribution. It enables collective decision-making and resource management among a heterogeneous workforce. 2) The second cornerstone is three kinds of operation modes supported by three types of workers, namely **autonomous, parallel, and expert/emergency modes**. Autonomous modes serve as the primary mode of operations, where activities are executed in an automated manner, mainly with the participation of digital and robotic humans. If an anomaly occurs, parallel modes are activated to enable remote access for biological humans, allowing them to work with other workers to resolve the problems. However, if the problem cannot be solved by parallel modes, emergency mode will be triggered, prompting the dispatch of related experts to directly address the situation. After the anomaly is solved, the operation mode will gradually cascade to autonomous modes. 3) The third cornerstone includes foundation models [27], scenarios engineering (SE) [28], and Human-Oriented Operating Systems (HOOS) [29]. Foundation models possess strong capabilities to solve various downstream sensing tasks, and can be recognized as the core of CSI. Scenarios engineering performs fine-tuning and validation of foundation models in specific scenarios to guarantee the interpretability and reliability of CSI. Furthermore, HOOS facilitates seamless communication and interaction between biological workers and digital or robotic workers, reducing laborious work and related physical and mental burdens of biological workers.

Promoted by these three cornerstones, CSI is now transitioning from algorithmic intelligence to linguistic intelligence, with the prosperity of Large Language Models (LLMs), such as ChatGPT [10], [30]. Natural language serves as a vital communication channel among diverse workforce in crowdsensing, and we believe that a novel 'conversational crowdsensing' paradigm represents an ideal form of linguistic intelligence for Industry 5.0, thereby facilitating the emergence of universally applicable and reliable foundational intelligence. In this survey, three levels of conversation during three sensing phases are devised to foster collaboration among three communities of workers in terms of the diverse sensing demands of various industrial activities across different scenarios, with the aim of achieving the '6S' goals of Safety, Security, Sustainability, Sensitivity, Smartness, and Services [31]–[33]. The core idea revolves around the design of LLM-based multi-agent systems with clearly defined roles and the integration of Human-AI cooperation technology (to enhance credibility and security).

Furthermore, we incorporate SE technologies to train and evaluate AI agents, aiming at enhancing the resilience and dependability of conversational crowdsensing.

In particular, the contributions of this paper can be summarized as follows:

1) Proposing conversational crowdsensing: we identify the limitations of current crowdsensing methods when applied in the context of Industry 5.0 and propose conversational crowdsensing, a PI-powered sensing approach that represents a specific form of crowdsensing in the stage of linguistic intelligence. By leveraging cutting-edge technologies such as DAO, LLMs, SE, and HOOS, three types of participants collaborate and coordinate in conversational crowdsensing to form a more autonomous and intelligent sensing pattern through natural language.

2) Devising the architecture: we design a '3&3&3' architecture of conversational crowdsensing, including three communities, three conversation levels, and three sensing phases. Specific roles with their respective embodied functions or activities are dedicately devised and organized in Human, AI, and Robot communities. Through three levels of effective conversation using nature language (i.e., inter-human, human-AI, and inter-AI), diverse tasks including information exchange, data transmission, and problem report are cooperatively accomplished by diverse workforce in three sensing stages.

3) Investigating the implementation scheme of underpinning technologies: we investigate the scheme of three underpinning technologies of for the successful implementation of conversational crowdsensing, including LLM-based multi-agent systems, conversational human-AI cooperation, and scenarios engineering. Regarding LLM-based multi-agent systems, we design a general framework for LLM-based conversable agent and introduce the DAO-based autonomous workflow control. The scenarios engineering approach based on parallel intelligence is designed to facilitate the acquisition of task completion capabilities by AI agents through fine-tuning. Moreover, we explore organization forms, interaction modes, and conversation designs of conversational human-AI cooperation to ensure the better cooperation between human participants and AI participants.

4) Prospecting future applications: we reveal the potential applications of conversational crowdsensing in various domains during the Industry 5.0 era, such as smarting mining, disaster prediction, and personal health monitoring, and provide the prototype implementation scheme.

## II. PRELIMINARY

In this section, we discuss the preliminary works from three aspects: (1) burgeoning parallel intelligence, (2) parallel intelligence powered crowdsensing, and (3) conversational crowdsourcing.

### A. Burgeoning Parallel Intelligence

The emergence of parallel intelligence is accompanied by the transition from traditional Cyber-Physical Systems (CPS)

to Cyber-Physical-Social Systems (CPSS) [34]. In comparison to CPS, CPSS involves the additional integration of diverse and complex human and social behaviors, posing challenges in building accurate models of actual systems, which is also known as the cognitive gap [35]. In 2004, Professor Fei-Yue Wang proposed the parallel system method [36] to describe and operate various entities in CPSS, of which the core is the ACP approach encompassing Artificial systems, Computational experiments, and Parallel execution [37]. This pioneering work delved into the realm of parallel intelligence, where parallel systems relate to composite systems comprising an actual system and one or more corresponding virtual artificial systems. With the advancement of artificial intelligence, the parallel system method continues to evolve towards intelligence, leading to the formal introduction of the concept of parallel intelligence in 2016 [5]. This concept has become the objective of parallel system methods, which aims to establish a cycle of data, knowledge, and action between the actual and artificial systems [38]. This cycle is facilitated by descriptive, predictive, and prescriptive intelligence, which correspond to the three steps of the ACP approach and represent their higher-level abstractions. Parallel intelligence enables the seamless feedback and interaction between the actual and artificial systems, offering significant potential for addressing challenges in modeling, analysis, management, and control within CPSS [39]. Currently, parallel intelligence has already been successfully applied in various domains, such as industry [40], [41], transportation [42]–[44], agriculture [45], sensing systems [19], [46], [47], and even artistic creation [48]. Most importantly, with the continuous development of concepts and technologies such as DAO [30], [49]–[51], Metaverse [24], [31] and foundation models [15], [21], [52]–[54], the connotation of parallel intelligence would undergo continuous evolution and updates.

### B. Parallel Intelligence powered Crowdsensing

Crowdsensing, as an emerging sensing paradigm, leverages the collective intelligence of individuals and organizations to acquire data for addressing urban-scale monitoring needs [12], [13], [55]. However, the expansion of complex sensing campaigns has brought forth a multitude of challenges for crowdsensing, such as substantial human efforts, inflexible organizational structure, sluggish system response, and limited application popularization. In this context, a novel generation of crowdsensing is required for enabling trustworthy, intelligent, interactive, collaborative, and autonomous sensing & computing [19]. The recent advancements give rise to the parallel intelligence powered crowdsensing, namely Crowdsensing Intelligence (CSI) [17]. By organizing Distributed/Decentralized Hybrid Workshops on Crowdsensing Intelligence (DHW-CSI), we aim to gain preliminary insights into its framework, participants, implementation methods, and development stages. The previous workshops on CSI have primarily focused on the DAO-based architecture [25], [26] and fundamental components of CSI [18], such as participants, methods, and stages. In the latest DHW-CSI, a concrete form of CSI, namely autonomous crowdsensing (ACS), was explored [56]. ACS organizes both professional sensing resources (e.g., satellites, IoT nodes, smart vehicles) and non-professional sensing resources (e.g., humans with smart devices) in the form of the DAO with the goal of sensing automation. By leveraging the comprehensive capabilities of LLMs [57], [58], digital and robotic participants can largely replace human labor in sensing tasks through a fresh 'conversational' mode [20]–[23], thereby minimizing reliance on human involvement. This would facilitate ACS to quickly respond to diverse sensing data demands and lower the cognitive and execution barriers for human participants. In addition, the adoption of cutting-edge technologies like scenarios engineering [28], [59]–[63], federated intelligence [64], and human-oriented operating systems [29] also contributes to the transition from traditional crowdsensing to ACS. With the progressive advancement of AI agent technology [65], this trend is increasingly becoming feasible.

### C. Conversational Crowdsourcing

The concept of conversational crowdsourcing originated in prior research that combined crowdsourcing and conversational agents [66] to achieve various goals, such as facilitating human-computer cooperative work [67], [68], training dialogue managers [69], improving user experiences [70], and acquiring knowledge from humans [71].

Conversational crowdsourcing was formally proposed in 2020 [72]. It features a conversational agent, which is capable of giving task instructions and asking questions by conversing with crowd workers, on the crowdsourcing task execution interface. A series of studies have been conducted to investigate different aspects of conversational crowdsourcing [73]–[76]. For instance, previous work on human-computer interaction focused on the effectiveness of using conversational crowdsourcing to improve user engagement and retention [73]. The findings showed that online participants who completed crowdsourcing tasks with a conversational agent were significantly more engaged and stayed significantly longer on task execution. An extended study of this work further showed that giving agents different conversational styles could affect the quality of crowdsourcing outcomes [74]. More recent works looked into visual representations and metaphorical representations of conversational agents (and their AI), and attempted to find more effective, efficient, healthy, and sustainable relationships between humans and AI in conversational crowdsourcing [75], [76].

In recent years, conversational crowdsourcing has been serving a variety of applications such as wellbeing check, mental therapy, building trust in decision support systems, enhancing human memorability in information retrieval [77]–[80]. Although crowdsourcing is considered to be an important domain application of general crowdsourcing, effectively enabling conversations in crowdsensing activities is still an unexplored research gap.

## III. ARCHITECTURE OF CONVERSATIONAL CROWDSENSING

The proposed conversational crowdsensing architecture, as depicted in Fig. 1, encompasses a range of roles with their respective embodied functions or activities. These roles are organized in human, AI, and Robot communities and engage in information exchange, data transmission, and communication using nature language in the conversational manner to cooperatively complete crowdsensing campaigns. Specifically, the architecture includes three levels of conversations: Inter-human, Human-AI, and Inter-AI conversations, as well as three phases of the entire campaign: Requesting, Scheduling, and Executing. In this section, we mainly present the composition of three communities and three levels of conversations. In the next section, we discuss the specific details of the three phases of conversational crowdsensing campaign.

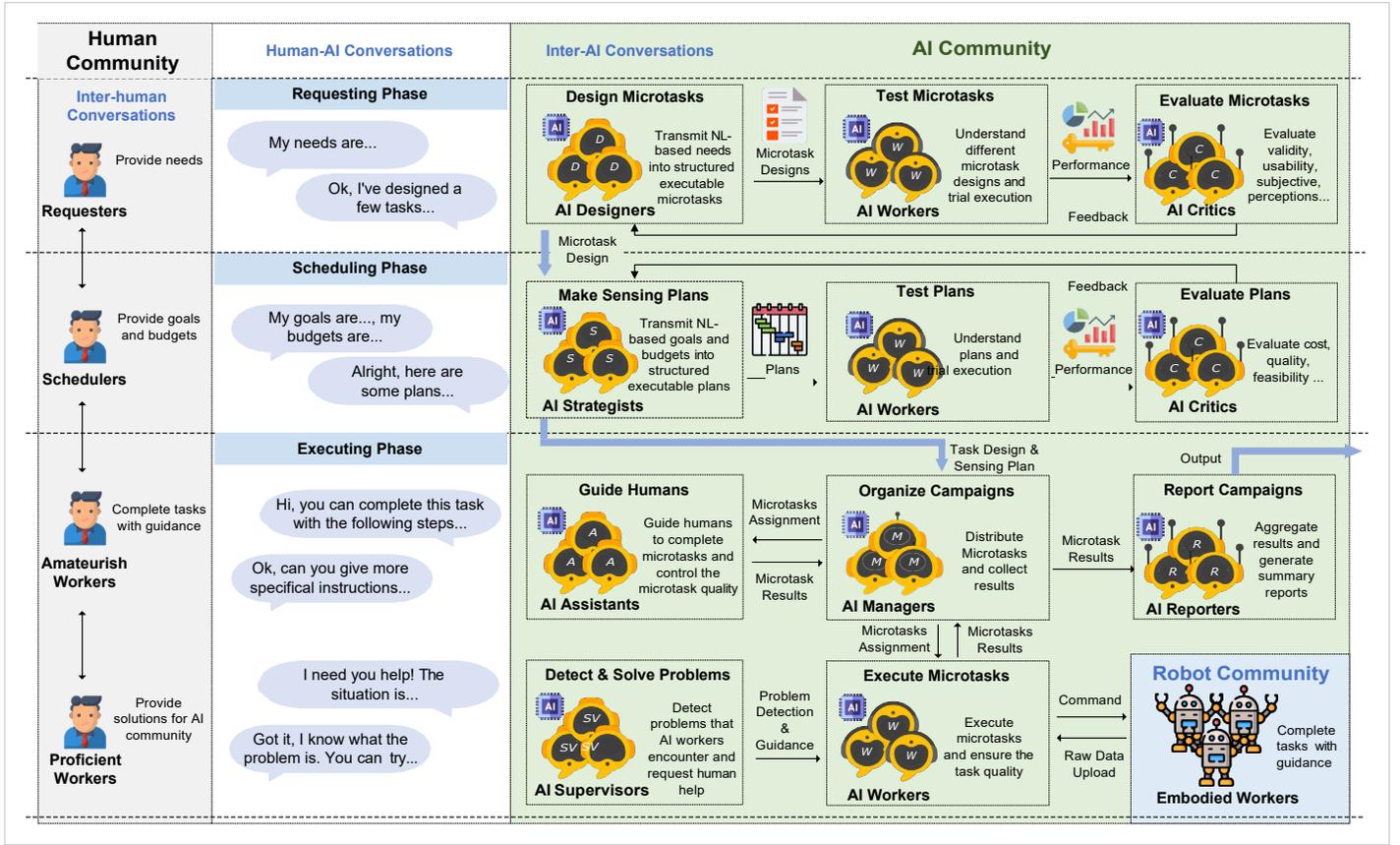

Fig. 1. The architecture of conversational crowdsensing.

### A. Three Communities in Conversational Crowdsensing

The elements involved in crowdsensing campaigns encompass humans, sensing vehicles, IoT sensing nodes, as well as diverse algorithms and software. Traditional crowdsensing typically operates within a human-centric architecture where humans organize and participant in major activities such as microtask design and sensing plan development. Due to the limited capacity of human beings, this traditional architecture hampers the response speed of sensing activities and imposes high demands on human expertise and skills. Therefore, by leveraging LLMs-based AI agents as the core, we propose organizing conversational crowdsensing to alleviate human workload and professional requirements, thereby enhancing the level of automation in crowdsensing campaigns. Specifi-cally, we reorganize elements of crowdsensing into the human, AI, and Robot communities.

*1) Human community:* The human community serves as the earliest and fundamental entity in crowdsensing campaigns. Human beings not only initiate and derive benefits from these campaigns, but also actively participate as workers. Distinguishing itself from traditional crowdsensing frameworks that solely consist of requesters and workers, we primarily introduce four distinct roles within the human community. **Requesters** are responsible for providing needs to drive the campaign toward a specific purpose. Different human needs require diverse task designs to respond, encompassing essential resources, microtask descriptions, and so forth. **Schedulers** are responsible for providing specific campaign goals and budgets to facilitate the determination of sensing plans. **Amateurish workers** can complete microtasks with the help of AI agents and submit the results, without necessitating any specialized expertise or capabilities. Their sole requirement is to employ their unique human capacity for recognition, reasoning, and device operation. In a city-centric sensing scenario, the widespread presence amateurish workers render them invaluable practitioners of microtasks. **Proficient workers** are highly skilled workers with specialized knowledge and expertise, capable of resolving intricate problems that cannot be addressed by AI. It should be noted that individuals in the human community can assume multiple roles simultaneously, such as being both a requester and a worker. Besides, with the assistance of AI agents, amateurish workers can quickly

TABLE I
THE EIGHT ROLES OF AI AGENTS IN CONVERSATIONAL CROWDSENSING ARE PRESENTED, ALONG WITH THEIR RESPECTIVE TASKS AND THE FUNDAMENTAL STEPS REQUIRED TO ACCOMPLISH EACH TASK. THE SYMBOL '*/#' MEANS THEY MUST ENGAGE IN CONVERSATION WITH HUMANS/AI AGENTS TO COMPLETE THE STEP.

| Roles | Tasks | Steps | Conversation with humans/AI agents (*/#) |
|-------|-------|-------|------------------------------------------|
| AI Designer | Design microtasks | Receive needs | * |
|  |  | Design microtasks |  |
|  |  | Send microtask designs to AI workers | # |
|  |  | Receive feedback from AI critics | # |
|  |  | Improve microtask designs |  |
|  |  | Send microtask designs to requesters | * |
| AI Worker | Execute microtasks | Receive microtasks | # |
|  |  | Perform microtasks with instructions |  |
|  |  | Send results of microtask to AI critics/managers | # |
| AI Critic | Evaluate performance | Receive results from AI workers | # |
|  |  | Evaluate performance from different perspective |  |
|  |  | Discuss with other AI critics | # |
|  |  | Send report to AI designers/strategists | # |
| AI Strategist | Make sensing plans | Receive goals, budgets, and microtask design | */# |
|  |  | Determine microtask list and assign microtasks to workers |  |
|  |  | Send microtask to AI workers | # |
|  |  | Receive feedback from AI critics | # |
|  |  | Improve sensing plans |  |
|  |  | Send sensing plan to schedulers | * |
| AI Manager | Organize crowdsensing campaigns | Receive microtask design and sensing plan | */# |
|  |  | Distribute microtasks to AI workers/assistants | # |
|  |  | Receive results from AI workers | # |
|  |  | Control quality of results |  |
|  |  | Send results to AI reporters | # |
| AI Assistant | Giude humans to complete microtasks | Receive microtasks | # |
|  |  | Generate multiple executable steps with detailed instructions |  |
|  |  | Guide and monitor human behavior step by step | * |
|  |  | Send results of microtask to AI managers | * |
| AI Supervisor | Detect problems and request human's help | Receive and monitor behaviors of AI workers and detect problems | # |
|  |  | Discuss with human for the solutions | * |
|  |  | Send guidance to AI workers | # |
| AI Reporter | Report crowdsensing campaigns | Receive results of microtask | # |
|  |  | Aggregate results and generate summary reports |  |
|  |  | Send reports to requesters/AI managers | */# |

improve themselves to proficient workers.

*2) AI community:* The design of the AI community serves as the core for conversational crowdsensing, encompassing a diverse range of roles fulfilled by LLM-based agents to assist humans in accomplishing various tasks, such as designing microtasks and making sensing plans. Moreover, it facilitates the automation of campaign operations and executions. We primarily present eight roles of LLM-based agents, i.e., AI designers, workers, critics, strategists, assistants, managers, reporters, and supervisors, all of which correspond to a task and possess the capability to exchange messages for information transmission with other agents/humans, as shown in Table I. Some of these roles also have the ability to execute tools through code or function execution [81]. For example, AI Strategists can extract human goals and budgets during conversations with schedulers and convert them into structured optimization problems. Furthermore, they can leverage relevant tools to solve these problems and transform the results into executable sensing plans [82], [83]. AI worker is a universal role capable of executing tasks in virtual scenarios to acquire performance evaluations for the designed microtasks and plans, or in the physical world by collaborating with embodied workers within the robot community.

*3) Robot community:* The robot community plays a crucial role in conversational crowdsensing, enabling AI workers to significantly enhance their capabilities and expand their business scope through embodied workers. This allows the plans formulated by the AI community to be effectively and automatically implemented in the physical world. The distinction made here is between embodied workers and traditional robot workers: robot workers may possess a certain level of autonomous planning and collaboration capabilities, often enabled by embedded algorithms and software. As a contrast, embodied workers in conversational crowdsensing serve as physical actuators while being equipped with corresponding AI workers from the AI community to provide guidance for their behavior. This facilitates the development of unified and scalable AI and robot communities, despite the heterogeneity of robot types, interfaces, parameters, etc. Therefore, the co-operation between embodied workers remains fundamentally a collaboration among AI managers. The embodied workers can seek assistance from human experts (proficient workers) through AI workers when they encounter difficulties during task execution, or alternatively, they can opt for remote takeover by human experts, a mode commonly referred to as parallel mode.

*B. Three levels of Conversation*

The inter and intra communications among the three communities are conducted through conversational means, pri-marily encompassing Inter-human conversations, Human-AI conversations, and Inter-AI conversations at three distinct levels. It is worth noting that we have disregarded Inter-robot conversations and robot-human/AI conversations since robots fundamentally rely on AI workers to communicate with other entities.

*1) Inter-human conversations:* The inter-human conversations serve as the most rudimentary form of linguistic communication, playing a pivotal role in crowdsensing campaigns. Particularly for intricate tasks, human workers can directly engage with requesters to acquire precise instructions or requirements pertaining to the task at hand, thereby ensuring alignment of objectives between them. Inter-human conversations ensure the non-destructive transmission of information and prevents information deviations caused by automated task design and planning processes. The traditional frameworks for crowdsensing typically require the avoidance of inter-human communication, to protect human privacy and ensure that each worker independently uploads high-quality data without being influenced by others. In contrast, conversational crowdsensing facilitates efficient information exchange, effective collaboration, and idea sharing among humans, enabling them to tackle more complex tasks. In addition, previous studies have noticed unhealthy relationships between requesters and workers. Particularly, workers have to shoulder a lot of invisible labor without being paid [84]–[86]. A major cause is considered to be the lack of a dialogue platform/mechanism where requesters and workers can equally change ideas with each other and discuss affairs regarding either high-level regulations or specific tasks. Such a platform could facilitate the formation of worker unions for online crowdsensing marketplaces, thus protecting the rights of human participants and ensuring that human workers are not manipulated by AI [87].

*2) Human-AI conversations:* The current advancement of LLMs enhances the natural language comprehension and output capabilities of AI agents, facilitating seamless and continuous communication between humans and AI. The conversational interaction, as compared to traditional human-computer interaction interfaces, offers a more natural and intuitive experience. It diminishes the reliance on specific technical knowledge and operational skills, allowing humans to focus their efforts on expressing needs and problems while efficiently obtaining information through instant AI feedback. This presents an excellent opportunity for non-professionals to engage in crowdsensing campaigns. Moreover, personalized AI agent customizations enhance conversational alignment with individuals' language habits and expressions, thereby augmenting human engagement and acceptance [73], [74]. In conversational crowdsensing, humans engage in microtask design and sensing plan formulation through dialogue with AI agents while receiving guidance during the sensing & reporting phase. Moreover, while current LLM-based agents possess remarkable capabilities for reasoning, thinking, and utilizing tools, and are increasingly excelling in various automated tasks, it remains crucial to assess the risks and feasibility of AI agents independently accomplishing diverse tasks. Consequently, ensuring an effective and appropriate level of human supervision becomes exceedingly significant. Conversational crowdsensing provides an interface for human interventions in addressing challenges that AI agents are unable to tackle independently, primarily through engaging in human-AI conversations. In general, through conversations, humans and AI agents can achieve efficient information exchange and complementary capabilities, so as to complete complex tasks in crowdsensing campaigns.

*3) Inter-AI conversations:* Through the acquisition of extensive web knowledge, LLMs have demonstrated remarkable potential in achieving human-level intelligence [88]. This progress has inspired researchers to develop methods for facilitating cooperative interactions among multiple AI agents through conversation. The utilization of multiple LLM-based agents has been shown in prior research to facilitate divergent thinking, enhance factuality and reasoning, provide validation, and mitigate the risk of hallucination [81]. Specifically, multiple LLM-based agents can be customized with distinct roles and embedded with specific functional tools. They then engage in conversation for information exchange and the completion of atomic tasks, enabling the automated resolution of complex tasks. The conversation pattern between agents can be static. For instance, Chen et al. [89] devised a chat chain to facilitate the automation of software development involving multiple roles such as CEO, CTO, Programmer, and Tester. Each role collaborates through dialogue to sequentially accomplish discrete tasks including Designing, Coding, Testing, and Documenting. The conversation mode can also be dynamic. For instance, Wu et al. [81] developed AutoGen, a multi-agent cooperative framework that incorporates the GroupChatManager module to facilitate complex and dynamic group conversations with dynamically selected speakers. These studies typically involve conducting multi-turn inter-AI conversations until termination conditions are met. Inter-AI conversations, in general, enable the automated execution of tasks or processes that would otherwise necessitate human efforts to connect them together. In conversational crowdsensing, we leverage inter-AI conversations to automate task design, plan creation, task assignment, and other related activities, thereby significantly enhancing the speed at which requesters' needs are addressed.

## IV. THREE PHASES OF CONVERSATIONAL CROWDSENSING

The process of conversational crowdsensing consists of three phases: requesting, scheduling, and executing. The primary objective of the requesting phase is to transmit human natural language (NL) based needs into executable microtasks by determining their fundamental form and descriptions. The primary objective of the scheduling phase is to devise a sensing plan that adheres to the constraints imposed by the goals and budgets of the crowdsensing campaign, specifically involving the allocation of multiple microtasks among workers with diverse attributes such as capability, cost, and reputation. The primary objective of the executing phase is to organize workers in carrying out their respective microtasks as per the sensing plan, while also collecting results for generating comprehensive reports and providing feedback to requesters. This section mainly introduces the main process of conversational crowdsensing in three phases with the joint participation of humans, AI agents and robots.

### A. Requesting phase

In the requesting phase, requesters will express their needs to AI designers in a conversational manner, while AI designers promptly respond by providing viable and high-quality microtask designs for humans. The design of microtasks is a highly specialized task that was predominantly carried out by humans in the previous crowdsensing frameworks. Specifically, it entails creating a clear microtask description with sufficient information, typically combining a concise title with detailed instructions [90]. In general, the instructions should be clear and comprehensible, providing adequate guidance on the microtasks to be performed and the methods to be employed. High-quality microtask design has the potential to enhance worker engagement, task completion rates, and overall satisfaction [73]. However, the design of high-quality microtasks poses a significant challenge even for human designers due to their limited understanding of the diverse demographics and abilities of potential workers. To enhance the performance of AI designers, we have introduced two types of AI roles - AI workers and AI critics - to collaboratively accomplish microtask design tasks at this phase.

When receiving human needs, AI designers will initially utilize the understanding and generation capabilities of the LLMs to generate a few preliminary schemes for microtask designs. Subsequently, these diverse design schemes are dispatched to AI workers who attempt to execute microtasks within virtual scenarios based on the provided descriptions. These AI workers can be preconfigured with varied knowledge backgrounds and abilities in order to yield distinct performances. Subsequently, AI critics consolidate these performances and assess the microtask design. This process can be implemented through a group chat dialogue, where distinct AI critics can be assigned to focus on various aspects such as validity, usability, etc. Following deliberation, the AI critics provide feedback to the AI designers regarding the preliminary schemes in order to facilitate the generation of higher quality microtask designs. The microtask design undergoes iterative refinement through this process. Following multiple iterations, AI designers provide feedback to the requesters and determine whether further improvement are necessary based on human input.

The requesting phase fully leverages the understanding and generation capabilities of LLM-based agents, as well as their collaborative abilities to accomplish complex tasks. Additionally, virtual scenarios are provided for AI workers to efficiently and effectively evaluate microtask designs. These virtual scenarios can be constructed through scenario engineering to ensure the achievement of '6S' goals.

### B. Scheduling phase

The execution of crowdsensing campaigns typically adheres to the constraints imposed by goals and budgets. These goals encompass requirements for data quality, spatio-temporal scale, and more, while budgets primarily involve cost limitations. Consequently, during the scheduling phase, a sensing plan must be devised based on these objectives and financial considerations. This includes determining a list of micro- tasks, selecting suitable workers, and assigning the microtasks to them. In conversational crowdsensing, AI strategists are responsible for developing a comprehensive sensing plan. Similar to the requesting phase, in order to enhance the output quality of these AI strategists, we have equipped them with both AI workers and AI critics. The AI workers execute sensing plans in virtual scenarios, which are then evaluated by the AI critics. The feedback provided by the AI critics is utilized by the AI strategists to refine and improve the overall quality of their sensing plan.

The development of sensing plans is a crucial research direction in crowdsensing research, particularly focusing on the challenging issue of microtask assignment. This problem can typically be formulated as an optimization problem and is commonly known to be NP-hard. Recent research utilizing LLMs as optimizers for completing optimization tasks has provided us with viable approaches to constructing AI strategists [82]. Firstly, the task assignment problem can be formalized based on the LLMs' comprehension abilities, and subsequently corresponding tools such as the LLM-driven evolutionary algorithm can be employed to solve the problem [83]. Although current exploratory work is still in its preliminary stages, we are able to recognize its potential.

### C. Executing phase

The executing phase encompasses various AI roles, with AI managers serving as the linchpin of this phase by orchestrating the entire crowdsensing campaign. Specifically, upon receiving the sensing plan and microtask design, AI managers will undertake precise microtask allocation and collect the results for AI reporters. The execution of microtasks can be divided into three modes, namely autonomous, parallel, and emergency modes. These modes involve the joint participation of human, AI, and embodied workers [18]. In **autonomous mode**, the majority of microtasks will be executed by AI workers to facilitate sensing automation. AI workers are capable of accomplishing specific types of microtasks in cyberspace, such as network information collection, image labeling, document writing, and more. Furthermore, AI workers can direct embodied workers to complete microtasks in the physical space. In order to ensure the trustworthiness and security of AI workers' behaviors during the execution of microtasks, the conversational crowdsensing system employs AI supervisors to monitor the conduct of AI workers, enabling timely detection of any encountered issues. Upon identification of a problem, the **parallel mode** is activated, prompting AI supervisors to seek assistance from proficient human workers by explaining the encountered issue and providing potential solutions. T The AI supervisors acquired the solution through deliberation with proficient workers and directed the AI workers in resolving the issues. In cases where a problem persists unresolved in parallel mode, **emergency mode** will be activated, enabling microtasks to be assigned to AI assistants who can guide human workers in completing them. AI assistants will generate more comprehensive and tailored instructions based on microtask descriptions and the individual characteristics of workers. As a result, even individuals without specialized knowledge or abilities can successfully complete microtasks with the help of AI assistants.

The AI reporters have the capability to aggregate the results of microtasks and generate real-time summary reports during the execution of such tasks. These reports can be sent to AI managers for effective monitoring of ongoing sensing campaigns or to requesters for prompt response to their requirements.

## V. Underpinning Technologies

The successful implementation of conversational crowdsensing necessitates the incorporation of at least three underpinning technologies. Firstly, LLM-based multi-agent systems facilitates conversational crowdsensing to establish a high-quality AI community, encompassing the design of LLM-based agents and their interactions with one another. Secondly, scenarios engineering plays a pivotal role in training and evaluating AI agents, thereby bolstering the robustness and reliability of conversational crowdsensing. Finally, the integration of Human-AI cooperation technology is imperative to ensure timely human involvement in crowdsensing campaigns and enhance the credibility and security thereof.

### A. LLM-based Multi-agent Systems for Conversational Crowdsensing

The performance of LLM-based multi-agent systems has been demonstrated in various domains. On one hand, this is attributed to the enhanced capabilities provided by LLM for individual autonomous agents. On the other hand, it is due to the emergence of more powerful cooperation among multiple agents [91]. When constructing LLM-based multi-agent systems for conversational crowdsensing, two challenges need to be addressed: (1) How to design individual agents that can be reused, customized, and capable of engaging in conversations. (2) How to achieve a straightforward and modular collaborative workflow control among multiple agents for accomplishing complex tasks. Therefore, in this section, we first introduce the conversable agent as a general framework for AI agents involved in conversational crowdsensing. Secondly, we present the workflow control based on the DAO.

*1) The design of LLM-based conversable agent:* The previous work presents a comprehensive framework for LLM-based autonomous agents, encompassing four key modules: a profiling module, a memory module, a planning module, and an action module [88]. Building upon this foundation, we have devised a refined framework for conversable agents in the context of conversational crowdsensing, as depicted in Fig. 2. In addition to the fundamental four modules, the conversable agent incorporates a unified conversation interface that facilitates collaborative interactions through conversations.

- **Profiling Module.** The profiling module aims to indicate the profiles of agent roles, typically incorporated into prompts to influence LLM behaviors. Profiles generally encompass a description of role tasks and the fundamental steps required for their completion. Additionally, they encompass collaborative relationships between agents, such as determining which AI roles should the agent initiate a conversation with after completing a task. Agent profiles could also define agents' characterizations that affect user experiences. Typical characterizations include metaphorical representations and conversational styles. A suitable metaphorical representation (e.g., a god, a human, an animal, an object, etc.) and a suitable conversational style (e.g., high-involvement and high-considerateness) of the AI, can positively affect work outcomes, and more importantly, improve humans' subjective perceptions during conversations [73], [74], [76], [92]–[94]. We will explain the details in section V-C3.

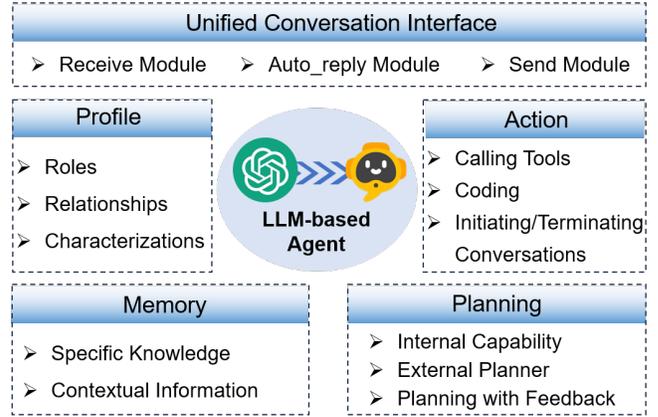

Fig. 2. The design of LLM-based conversable agent.

- **Memory Module.** The memory module facilitates the agent in accumulating experience, self-evolution, and enhancing its behavior to be more consistent, rational, and effective. Memory structures typically encompass long-term memory and short-term memory. Long-term memory is primarily associated with the agent's underlying business logic and encompasses specific domain knowledge and skills. Short-term memory serves as a repository for contextual information pertaining to the agent's ongoing task, such as current task progress or dialogue topic. Since the agent always interact with a continuous dynamic environment, the continuous actions show a high degree of correlation. Therefore, the capture of short-term memory is very important and usually cannot be ignored. Moreover, memories can be stored using various methods such as natural language, embeddings, databases, and structured lists.

- **Planning Module.** The planning module is designed to empower the agent with the capability to decompose a complex task into simpler subtasks and solve them individually. Some existing research endeavors leverage the internal capacity of LLMs for accomplishing intricate task planning, such as Chain of Thought (CoT), Zero-shot-CoT, Self-consistent CoT, Tree of Thoughts (ToT), among others [88]. Furthermore, agents can invoke external planners when confronted with domain-specific challenges. For instance, LLM+P has been developed to transform the task description into a formal Planning Domain Definition Languages (PDDL) [95]. In addition, when confronted with intricate sensing tasks in real-world scenarios, it is occasionally unfeasible to generate a high-quality scheme directly from the out-set. Therefore, the scheme can be gradually optimized through feedback obtained from either human or virtual environments. The planning with feedback method is applied in the requesting and scheduling stages.

- **Action Module.** The action modules enable agents to accomplish corresponding tasks and interact with other entities. Firstly, LLM-based agents need the capability to call external tools for expanding the range of actions available. These external tools typically encompass APIs, databases, knowledge bases, and models. Secondly, the agent primarily realizes complex interaction functions through two types of actions: coding and initiating/terminating conversations. In addition to natural language, code serves as a crucial communication medium between agents and an essential means for invoking external tools. To implement intricate workflows, agents must initiate conversations at the appropriate time to trigger interactions while also possessing the ability to explicitly terminate these conversations in order to advance the workflow.

- **Unified Conversation Interface.** The unified conversational interface empowers conversable agents to engage in conversation-centric interactions, encompassing a receive/send module for message reception/transmission and a generate auto reply module for executing actions and generating responses based on received messages. Consequently, the initial step for AI roles depicted in Table I to accomplish their respective tasks involves receiving messages, while the final step entails sending messages. The implementation of the unified conversational interface allows us to construct agent interactions in a cost-effective and modular manner, achieved through the definition of agent auto reply modules. Simultaneously, it facilitates the realization of autonomous workflow control based on the DAO.

*2) The workflow control based on the DAO:* The unified conversational interface of conversable agents enables us to leverage the agent's automatic reply mechanism for driving workflow control via dialogue flow, thereby achieving autonomous workflow control based on the DAO, as depicted in Fig. 3. Specifically, when an agent receives a message from another agent, it is automatically invoked to generate a reply and send the response message in the form of smart contracts. The dialogue flow is naturally induced once the agent's auto reply function is defined and the conversation is initialized under this mechanism, enabling seamless conversations between agents without any additional control modules. In order to achieve comprehensive distributed process control, we organize AI agents in the DAO-based communities, where the messages exchanged between them are securely stored on the blockchain through smart contracts. Such an organizational structure also enhances privacy protection.

*B. Capability Acquisition Based on Scenarios Engineering*

The current multi-agent systems based on LLMs have demonstrated impressive capabilities; however, they may encounter challenges when deployed in real-world scenarios. On one hand, this can be attributed to their lack of task-specific abilities, skills, and experience. On the other hand, LLM-based agents are susceptible to generating hallucinations that result in inaccurate or even detrimental outputs. Therefore, in addition to designing LLM-based conversable agent, we also need to focus on enhancing the agents' capabilities to qualify them for completing various tasks on behalf of humans. This section initially introduces several existing methods for acquiring capabilities and subsequently presents scenarios engineering based on parallel intelligence to enhance agents' competence.

*1) Capability acquisition methods for AI agent:* The current methods for acquiring capabilities in AI agents can be classified into two categories based on whether they require fine-tuning [88]. Prompt engineering is a capability acquisition method that does not require fine-tuning. In prompt engineering, valuable information must be incorporated into the prompt to enhance model functionality or release existing model functionality. The efficiency of prompt engineering can be influenced by prompt patterns, which offer reusable solutions to specific problems and standardized methods for customizing AI agent output and interaction in the field of prompt engineering. White et al. [96] summarized similar formats of prompt patterns and gave 16 common prompt patterns, including Meta Language Creation, Output Automater, etc. In addition to prompt engineering, researchers have devised specialized mechanisms to augment the capabilities of AI agents, such as Trial-and-error, Crowd-sourcing, Experience Accumulation, and Self-driven Evolution [88]. The utilization of multi-intelligence cooperation in this paper also represents a distinctive mechanism.

Another approach to enhance an agent's task completion capability is by directly fine-tuning the agent using task-specific datasets. For instance, Mind2Web offers a collection of over 2,000 data samples from 137 websites across 31 domains for constructing generalist AI agents for the web [97]. The availability of real-world datasets may sometimes be limited, necessitating the use of generated data as an alternative for fine-tuning. For example, Liu et al. [98] established a sandbox environment to generate and collect 169k samples of agent social interactions, which could be utilized to refine the agent's behavior in accordance with established societal values.

For the professional business like conversational crowd-sensing, fine-tuning is essential to ensure the availability and reliability of the entire process. Conversational crowdsensing includes eight roles of LLM-based agent, each with their own tasks that are interconnected. For instance, AI managers should possess an understanding of how AI workers perform microtasks to effectively organize the overall crowdsensing campaign. However, most of the existing datasets in the field of crowdsensing are task-oriented, and it is difficult to find all the data needed for research in a single data set [18].

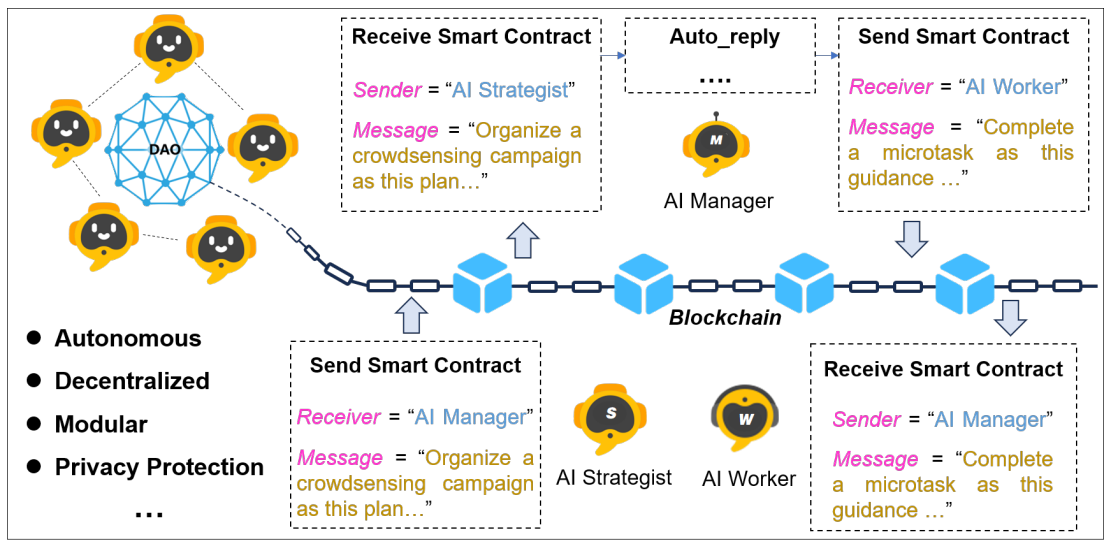

Fig. 3. The autonomous workflow control based on the DAO.

Furthermore, these datasets are static and fixed, failing to capture the uncertainty and dynamics present in real crowdsensing scenarios. These factors present formidable obstacles when it comes to fine-tuning AI agents in conversational crowdsensing. To address these challenges, we designed the scenarios engineering approach based on parallel intelligence to generate, storage, process, and index the data for the fine-tuning.

*2) Scenarios engineering based on parallel intelligence:*
A scenario is a coherent, internally consistent, and plausible depiction of a potential future state of the world. Scenario data generally comprehensively encompasses numerous activities within a specific space-time range and their corresponding impacts. Scenario data represents a more comprehensive and rigorous form of information compared to task-specific data [60]. The fine-tuning process, based on scenario data, enhances the stability and reliability of LLM-based agents by incorporating dynamic and uncertain factors that are present in real-world scenarios but often overlooked during training. In addition, utilizing relevant scenario data for fine-tuning AI agents operating within the same business process can enhance their collaborative capabilities, thereby optimizing overall system performance. Scenario engineering involves leveraging a diverse range of scenario data to train and shape AI systems in order to achieve the '6S' goals.

We design the scenario engineering approach based on parallel intelligence to fine-tune AI agents in conversational crowdsensing, as shown in Fig. 4. Due to the inherent sparsity and inadequacy of real scenario datasets, we recommend leveraging data and models to construct artificial systems that accurately represent their real counterparts, known as descriptive intelligence. The scenario data in crowdsensing primarily consists of four essential components: participants, environment, events, and platforms. Ren et al. [99] proposed an artificial system for vehicular crowdsensing and extensively discussed these four aspects of data. Artificial systems can generate a substantial volume of scenario data for the fine-tuning of AI agent, which is the so-called predictive intelligence. The content of scenario data may vary among different agents, but they are all generated based on a unified logic and interconnected. For instance, the crowd-mapping approach is specifically designed to generate crowdsourced object annotations from street-level imagery, requiring the assignment of microtasks to workers and guiding workers to complete these microtasks [100]. We can naturally adopt AI agents to accomplish these two tasks. For AI worker-oriented scenarios, they may present microtask execution scenarios in the form of images, 3D models, etc. For AI manager-oriented scenarios, they may include more diverse assignment modes data, such as single-queue or multi-queue, as shown in Fig. 4. The well-tuned agents can achieve superior performance in actual systems, thereby facilitating the infusion of higher quality data into artificial systems, which is known as prescriptive intelligence. By employing scenario engineering based on parallel intelligence, agents are capable of generating high-quality behaviors and continuously evolving in novel scenarios to adapt to dynamic real-world business scenarios.

### C. Conversational Human-AI Cooperation

The goal of conversational human-AI cooperation is to achieve a better understanding between human participants and AI participants, and therefore, to improve the outcome quality, execution efficiency, user experience, and so on [101]. We discuss organization forms, interaction modes, and conversation designs of conversational human-AI cooperation, as shown in Fig. 5.

*1) Organization Forms:* There are various organization forms to serve different crowdsensing goals and needs. In this work, we explain three typical forms – independent conversations, conversation chains, and hierarchical discussions – that we consider to be prevalent in existing crowdsensing campaigns or to be of good potential in the age of LLM. The organization of conversational human-AI cooperation is not limited to these three forms. It can be a mix of these forms or any other possible forms that can make human-AI cooperation work.

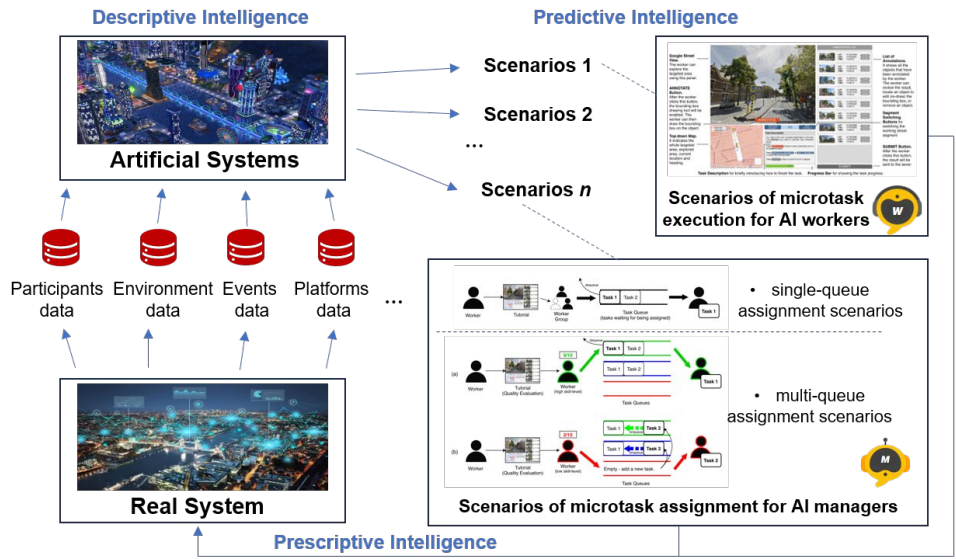

Fig. 4. Scenarios engineering based on parallel intelligence.

- **Independent Conversations.** This form refers to a traditional tree-like organization structure, letting workers/participants complete tasks independently. This form of organization ensures independent task execution, and therefore, avoids potential biases or cheating caused by unexpected communications between workers.

- **Conversation Chains.** This form refers to human-AI cooperation that is organized in a sequential chain-like way. A conversation chain enables human/AI workers to complete multi-step tasks satisfying some classical crowd work pipelines such as the find-fix-verify pattern [102], or to conduct multi-phase inference and reasoning achieving complex causal analysis.

- **Hierarchical Discussions.** Hierarchical discussion provides a possible human-AI cooperation form that enables inter-worker communications. Hierarchical discussion features a multi-layer structure of the organization, having both human and AI workers involved in the discussion. This is particularly suitable for complex tasks that require brainstorming and innovative ideas.

*2) Interaction Modes:* The organization provides a big picture of how humans and AI should cooperate in conversational crowdsensing. The cooperation inevitably leads to interaction between humans and AI. Depending on what roles humans or AI should play during cooperation, we introduce two main interaction modes – human-in-the-loop mode and AI-in-the-loop mode.

- **Human-in-the-Loop.** Human-in-the-loop means the whole process is dominated by AI and uses minimal support from humans. The process is supposed to be fully automated, but sometimes AI might have a problem or need humans to make a decision at some point in time. Humans play a role in augmenting AI. Human-in-the-loop is a concept proposed by AI researchers and practitioners, and is becoming increasingly popular in AI-related research. In this interaction mode, AI usually initiates the conversation when it detects a problem or needs a human decision at a certain stage. The conversation is task-oriented and specifically designed to let humans understand the situation and address the issue. The conversation flow should consider a human-in-the-loop cooperation pipeline consisting of three stages, i.e., problem detection, problem explanation, and problem solving [103].

- **AI-in-the-Loop.** AI-in-the-loop is another interaction mode that emphasizes human dominance. In this mode, humans take charge of the whole cooperation process, and occasionally consult AI for assistance. AI plays a role in augmenting humans. We argue that research in human-AI cooperation should particularly pay attention to AI-in-the-loop mode, not only due to some types of crowdsensing tasks (like survey studies) that humans will still play main roles in the future even if AI advances a lot, but also owing to the growing risk as humans gradually rely on AI decisions. In this interaction mode, humans should initiate the conversation when they encounter problems or troubles in terms of understanding task content, using devices, or reporting results.

*3) Conversation Designs:* No matter whether humans should augment AI or AI should augment humans, in the context of conversational crowdsensing, human-AI cooperation is preferred to be conducted through conversations. Conversation design is an important factor affecting crowdsensing output quality and user experience. We discuss conversation design in two aspects – conversational styles and metaphorical representations.

- **Conversational Styles.** A conversational style is a combination of different linguistic devices. Deborah Tannen identifies nine dimensions of linguistic devices that are closely related to conversational styles, namely personal focus of topic, paralinguistic features, enthusiasm, use of questions, pacing, use of repetition, topic cohesion, tolerance of silence, and laughter [92], [93]. Considering these linguistic devices altogether, the conversational style can be depicted using a spectrum with two ends – high-involvement and high-considerateness. Previous studies have shown that conversational styles can affect crowdsourcing in terms of user engagement and output quality, particularly when conversational styles of speakers are aligned [73], [74], [94]. To achieve better cooperation between humans and AI, future work could focus on quick style estimation and natural style alignment.

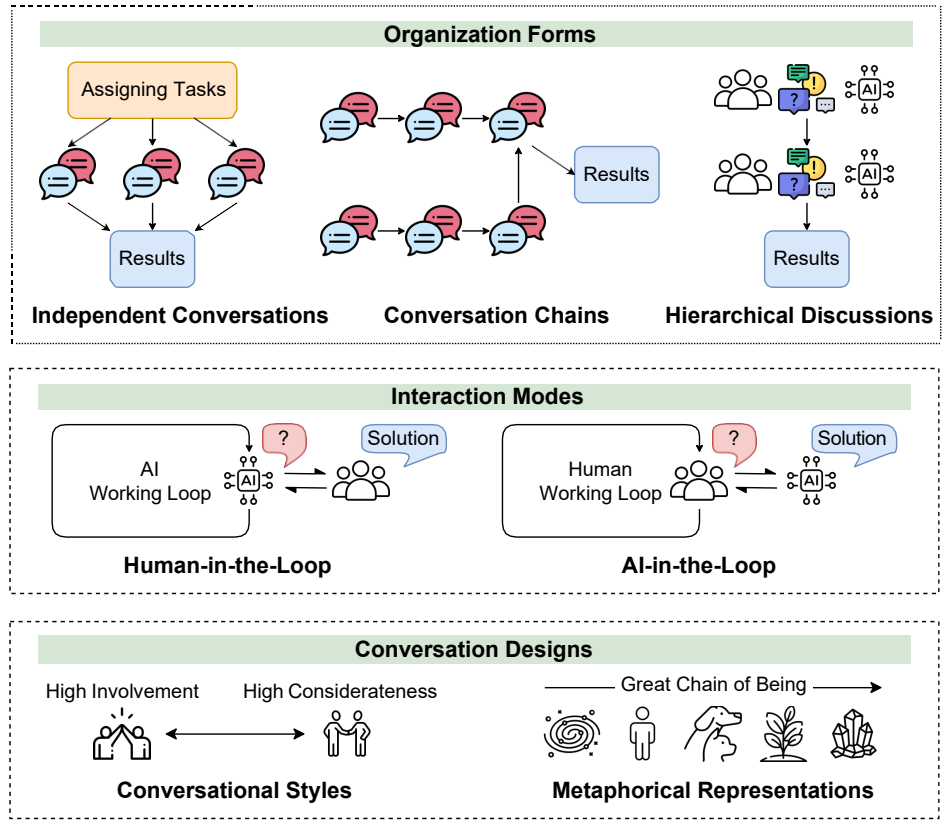

Fig. 5. Organization forms, interaction modes, and conversation designs of human-AI cooperation in conversational crowdsensing.

- **Metaphorical Representations.** To improve the quality of conversational human-AI cooperation, there is a need in terms of developing a mental model of AI agents, for humans involved in the conversation. This mental model usually refers to metaphorical representations [104], [105]. Recent research started to pay attention to the effect of metaphorical representation of AI agents in human-AI interactions. Previous work suggested that giving AI agents a human-level metaphor could affect users' subjective perceptions [105]. In addition, researchers also investigated the use of non-human AI metaphors, such as using animals (e.g., an owl in Duolingo). In the field of conversational crowdsourcing, the role of metaphorical representation was also explored using the 'Great Chain of Being' framework. Most prior works focused on simple human-AI cooperation situations, where AI only plays one role generally. The emergence of LLM leads to a rapid increase in AI roles. How to assign AI roles (as shown in Table I) suitable metaphorical representations needs further exploration.

## VI. APPLICATIONS

As illustrated in Fig. 6, the proposed conversational crowdsensing has been extensively explored in this work and can be easily applied across various industrial stages (e.g., design, perception, production, logistics, and user feedback) of Industry 5.0 in diverse fields, such as mining, manufacturing, environmental monitoring, and healthcare. Consequently, some of the existing and potential future case studies are presented in this section.

### A. Urban Dynamics

The dynamic monitoring of various systems' operating states in a city is crucial for effective urban governance, such as environmental monitoring, traffic management, and ecological protection. Given the significance of data and information in urban governance, crowdsensing and spatial crowdsensing hold promise as a promising solution to improve urban operational efficiency and enhance urban resilience. This section presents a case study that applies conversational crowdsensing to disaster prediction and management within the context of Industry 5.0.

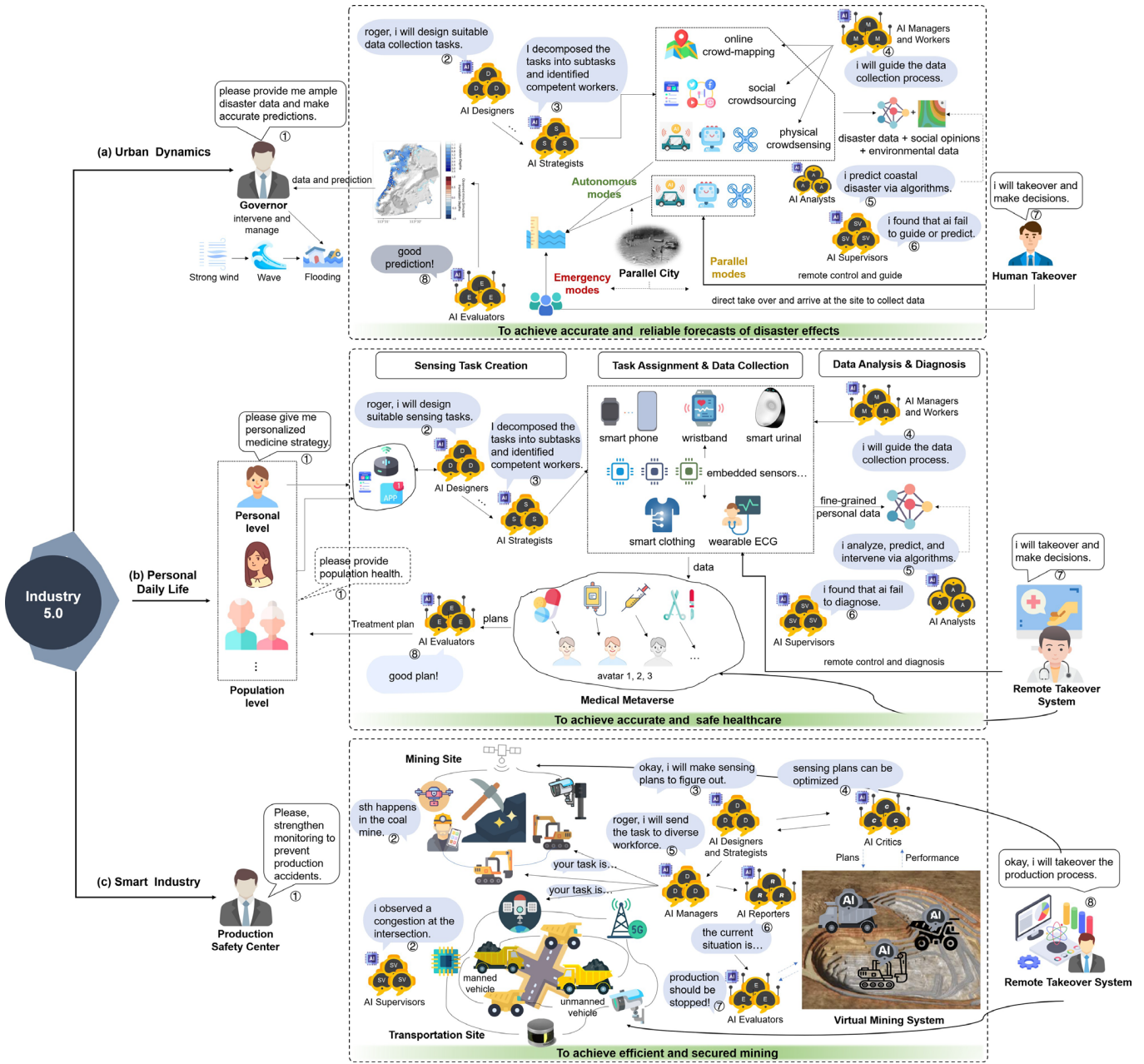

Fig. 6. Potential applications of conversational crowdsensing in the era of Industry 5.0.

The conversational crowdsensing approach will build a human-oriented operation system and provide smart services for disaster prediction and management tasks by developing human-machine interaction interfaces and applying multi-modal large models. Digital workers, robotic workers, and biological workers will be included and coordinated within this operation system for different scenarios under the mixture of Autonomous Modes (AM), Parallel Modes (PM), and Expert/Emergency Modes (EM) [106]. As described in Fig. 6(a), these modes play various roles in the workflow of this example. The primary mode of operations is AM, which accounts for over 80% of cases and mainly involves digital and robotic workers. The implementation of LLMs allows for the replacement or assistance of humans in various activities, thereby facilitating the automation of crowdsensing campaigns. AI workers take on a variety of roles in the sensing campaign to realize autonomous task clarity, task design, plan scheduling, task execution, and more. PM would be activated in fewer than 15% of cases, providing remote access for human experts to resolve any unforeseen issues or failures that may arise during the process of data collection and data analysis. In this scenario, AI supervisors effectively convey the failure situation to human operators through natural language. If an issue persists even after PM deployment, the corresponding data collection process will transition to EM, which accounts for less than 5% of cases. In such instances, human experts or emergency teams intervene directly at the site to resolve the problem. Once the issue is resolved, the data collection process will transition back to AM after being remotely monitored for a specified duration in PM. In this manner, accurate and reliable forecasts of disaster effects can be achieved.

### B. Personal Daily Life

With the evolving demands of human life-and-study, the traditional technologies targeting for personal daily life have also undergone a transformation. Enabled by the paradigm of Industry 4.0, social media recommendation systems, remote education platforms, and personal health monitoring technologies have been realized. Undoubtedly, conversational crowdsensing will play a pivotal role in personal daily life of Industry 5.0 era, aiming to achieve a more smart, customizable, interactive, and reliable life-and-study mode. As depicted in Fig. 6(b), we take personalized health monitoring as an example to illustrate this point.

It is evident that AI and robotic workers accomplish the majority of subtasks in a cooperative and autonomous manner, while biological human workers primarily engage in creative and instructional tasks. During the stage of **sensing task creation**, a biological human would request a personalized medical strategy from the AI Designer through dialogue. After receiving the instruction, AI designer will endeavor to comprehend the human request's description and design appropriate sensing tasks. Then, AI strategists would identify currently idle and available robotic facilities, assess their capabilities, and decompose the sensing task into subtasks for allocation. By utilizing the available information of devices and subtasks, AI managers and workers can efficiently assign tasks by invoking tools or functional interfaces and guide different robotic workers to cooperate to accomplish data collection through linguistic instructions during the stage of **task assignment and data collection**. After the collection of fine-grained personal data, AI analysts will employ deep learning algorithms for analysis and prediction. In case of an unsuccessful AI diagnosis, AI supervisor will communicate the situation to a human doctor. Subsequently, the doctor remotely takes over and utilizes his/her expertise and experience to make a diagnosis. Furthermore, in the future, personalized medicine strategies can be optimized within the medical metaverse by conducting experiments on diverse avatars. After optimizing individual treatment effects through conversational crowdsensing, the promotion of health within a specific target population can also be accomplished [107].

### C. Smart Industry

In the mining industry, remarkable transformations have occurred in the production mode due to the implementation of advanced technologies such as parallel intelligence, unmanned driving, LLMs, blockchain, and autonomous sensing. In recent years, parallel end-to-end autonomous mining [108], [109] and meta-mining [110] have been proposed as potential solutions to the intractable issues (e.g., extensive human involvement, limited flexibility, ineffective human-computer interaction, poor safety, and reduced mining efficiency) faced by traditional mining systems, aiming to achieve a more intelligent, customizable, secure, efficient, and sustainable mining mode. In general, the mining task includes several subtasks such as mining, loading, transportation, etc., with each subtask necessitating the support of advanced perception techniques. As shown in Fig. 6(c), we concisely demonstrate how our proposed conversational crowdsensing working in mining and transportation activities.

During the mining process, human operators may identify abnormal conditions and communicate them to AI systems through language interaction. AI designers and managers will then design and assign specific sensing tasks to various industrial IoT sensors for data collection. The collected data is utilized in a virtual mining system for super-real-time computational experiments, enabling timely assessment of the feasibility of the current mining plan. Subsequently, an AI evaluator assesses the plan and alerts humans to cease dangerous mining activities promptly. In this scenario, effective dialogue and communication among three kinds of participants ensure safe production.

When multiple vehicles (e.g., manned and unmanned vehicles) work together to accomplish transportation tasks, AI supervisors would promptly notify a remote human operator in the event of congestion at a specific intersection caused by harsh environmental conditions or communication disruptions. The human operator would then take over some of the unmanned vehicles and instruct AI workers to design sensing tasks through language communications to enhance the current situational awareness. Subsequently, AI workers will deploy various facilities and sensors from roadside infrastructure and vehicles, guide data collection process to enrich the perception field, and ultimately enhance efficiency, stability and robustness of the transportation system under the cooperation condition of AI assistance and human takeover.

### VII. Conclusions

In this survey, a parallel intelligence-powered sensing approach, namely conversational crowdsensing, is proposed, providing a new perspective on the design and development of a sensing paradigm for Industry 5.0. With the preliminary works introduced at first, we primarily designed the basic architecture of conversational crowdsensing, encompassing three communities of sensing workers, three phases in workflow (i.e., requesting, scheduling, and executing), and three levels of conversation involved in these phases (i.e., inter-human conversations, human-AI conversations, and inter-AI conversations). Furthermore, we introduced the underpinning technologies for implementing conversational crowdsensing and presented potential applications of conversational crowdsensing (e.g., smart mining). In the future, the emergence of conversational crowdsensing is anticipated to facilitate the development of universally applicable and reliable foundational intelligence, thereby fostering its widespread adoption across diverse industrial activities in Industry 5.0.

### ABOUT THE AUTHOR

**Zhengqiu Zhu** received the Ph.D. degree in management science and engineering from National University of Defense Technology, Changsha, China, in 2023. He is currently a lecturer with the College of Systems Engineering, National University of Defense Technology, Changsha, China. His research interests include mobile crowdsensing, social computing, and LLM-based AI agent.

**Yong Zhao** received the M.E. degree in control science and engineering from the National University of Defense Technology, Changsha, China, in 2021, where he is currently working toward the Ph.D. degree in management science and engineering. His current research focuses on crowdsensing, human-AI interaction, and intelligent transportation systems.

**Bin Chen** received the Ph.D. degree in control science and engineering from the National University of Defense Technology, Changsha, China, in 2010. He is currently an Associate Research Fellow with the College of Systems Engineering, National University of Defense Technology, Changsha, China. His research focuses on parallel simulation, mobile crowdsensing, and micro-task crowdsourcing.

**Sihang Qiu** received the Ph.D. degree from the Web Information Systems Group, TU Delft, Delft, The Netherlands, in 2021. He is currently an Associate Professor with the College of Systems Engineering, National University of Defense Technology, Changsha, China. His research focuses on human-centered AI, crowd computing, and conversational agents. Detailed Bio of Sihang Qiu can be found at https://sihangqiu.com/.

**Kai Xu** received the Ph.D. degree in control science and engineering from the National University of Defense Technology, Changsha, China, in 2020. He is currently a lecturer with the College of Systems Engineering, National University of Defense Technology, Changsha, China. His research interests include modeling and simulation of complex systems, artificial intelligence, and cognitive modeling.

**Quanjun Yin** is a professor of the National University of Defense Technology, Changsha, Hunan, China. His research interests include cooperation and negotiation, cloud-based simulation, and edge computing.

**Jincai Huang** is a Professor of the National University of Defense Technology, Changsha, Hunan, China. His main research interests include artificial general intelligence, deep reinforcement learning, and multi-agent systems.

**Zhong Liu** is a professor of the National University of Defense Technology, Changsha, Hunan, China. He is also the head of Science and Technology Innovation Team, Ministry of Education. His main research interests include artificial general intelligence, deep reinforcement learning, and multi-agent systems.

**Fei-Yue Wang (Fellow, IEEE)** received the Ph.D. degree in computer and systems engineering from the Rensselaer Polytechnic Institute, USA, in 1990. He is the State Specially Appointed Expert and the Director of the State Key Laboratory for Management and Control of Complex Systems. His current research focuses on methods and applications for parallel systems, social computing, parallel intelligence, and knowledge automation. Detailed Bio of Fei-Yue Wang can be found at www.compsys.ia.ac.cn/people/wangfeiyue.html.